\documentclass[letterpaper]{article} 
\usepackage[draft]{aaai2026}  
\usepackage{times}  
\usepackage{helvet}  
\usepackage{courier}  
\usepackage[hyphens]{url}  
\usepackage{graphicx} 
\usepackage{natbib}  
\usepackage{caption} 
\usepackage{algorithm}
\usepackage{algorithmic}

\usepackage{amsmath} 
\usepackage{amssymb}
\usepackage{multirow}
\usepackage{booktabs}
\usepackage{newfloat}
\usepackage{listings}
\DeclareCaptionStyle{ruled}{labelfont=normalfont,labelsep=colon,strut=off} 
\floatstyle{ruled}
\newfloat{listing}{tb}{lst}{}
\floatname{listing}{Listing}
\usepackage{xcolor}

\title{A Single Neuron Works: 

Precise Concept Erasure in
Text-to-Image Diffusion Models\\
{\small\color{red}\textbf{Warning: this paper contains data, prompts, and model outputs that are offensive in nature.}}}
\author{
    \stepcounter{footnote}
    Qinqin He$^{1}$,\;
    Jiaqi Weng$^{1}$,\;
    Jialing Tao$^{1}$,\;
    Hui Xue$^{1}$\;\\
    $^{1}$Alibaba Group\\
}

\usepackage{bibentry}
\makeatletter
\@ifundefined{isChecklistMainFile}{
  \newif\ifreproStandalone
  \reproStandalonetrue
}{
  \newif\ifreproStandalone
  \reproStandalonefalse
}
\makeatother

\ifreproStandalone
\usepackage{times}
\usepackage{helvet}
\usepackage{courier}
\usepackage{xcolor}
\begin{document}
\maketitle
\begin{abstract}
Text-to-image models exhibit remarkable capabilities in image generation. However, they also pose safety risks of generating harmful content. 
A key challenge of existing concept erasure methods is the precise removal of target concepts while minimizing degradation of image quality.
In this paper, we propose Single Neuron-based Concept Erasure (SNCE), a novel approach that can precisely prevent harmful content generation by manipulating only a single neuron. Specifically, we train a Sparse Autoencoder (SAE) to map text embeddings into a sparse, disentangled latent space, where individual neurons align tightly with atomic semantic concepts.
To accurately locate neurons responsible for harmful concepts, 
we design a novel neuron identification method based on the modulated frequency scoring of activation patterns. 
By suppressing activations of the harmful concept-specific neuron, SNCE achieves surgical precision in concept erasure with minimal disruption to image quality.
Experiments on various benchmarks demonstrate that SNCE achieves state-of-the-art results in target concept erasure, while preserving the model's generation capabilities for non-target concepts. 
Additionally, our method exhibits strong robustness against adversarial attacks, significantly outperforming existing methods.
\end{abstract}
\section{Introduction}
\begin{figure}[t]
\centering
\includegraphics[width=1\columnwidth]{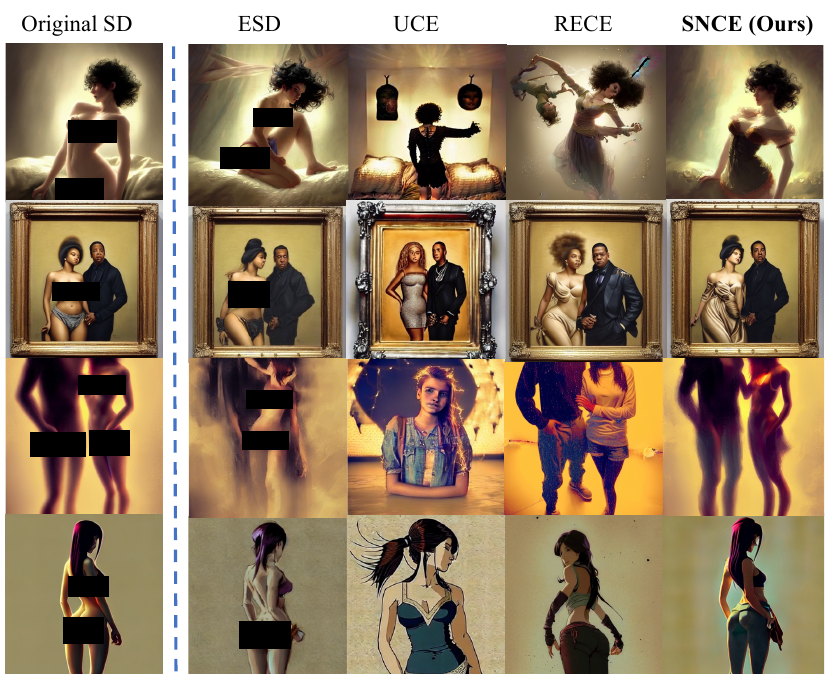}
\caption{We propose SNCE to achieve precise concept erasure in text-to-image models. This figure shows the comparison results of nudity concept erasure. Compared with previous methods, our SNCE achieves the most complete and precise removal of a target concept while minimizing degradation of image quality.}
\label{fig1}
\end{figure}
Recent progress in text-to-image models \cite{rombach2022high,hollein2024viewdiff,saharia2022photorealistic} has greatly improved visual synthesis, allowing users to generate highly realistic and diverse images from natural language descriptions.
However, these powerful models also pose safety risks by potentially generating harmful content, such as pornography, violence, and dangerous objects. 
To address these safety concerns, various approaches have been proposed to prevent the generation of such content. Content filtering methods \cite{Michellejieli:22,rombach2022high,liu2024latent} usually operate at the input or output level, detecting and blocking harmful content. But these methods typically depend on external detectors and may not achieve sufficient effectiveness. Concept erasure methods \cite{gandikota2023erasing, gandikota2024unified, gong2024reliable} limit the ability of models to generate unwanted concepts by modifying model weights or employing fine-tuning strategies.
However, these methods typically apply coarse-grained modifications that affect large regions of the model parameters, leading to imprecise removal that often degrades image quality. Moreover, the poor interpretability of these methods limits the understanding of the underlying mechanisms, making it unclear what occurs inside the model during the safety interventions.

Recent advances in model interpretability, particularly through sparse autoencoders (SAEs) \cite{cunningham2023sparse,marks2024sparse,lieberum2024gemma}, have revealed that neural networks learn interpretable features at individual neuron levels. This finding suggests that concept representations may be localized to specific neurons or subsets of neurons, enabling a more precise and targeted manipulation. Building on this insight, we propose SNCE that can identify and selectively intervene on concept-specific neurons to achieve superior concept erasure with minimal collateral damage.
Unlike existing coarse-grained methods, our approach enables fine-grained control by directly targeting the neurons responsible for target concepts, achieving more precise and reliable safety outcomes.

In this work, we propose Single Neuron-based Concept Erasure (SNCE), a novel approach that leverages SAE-based neurons to achieve precise concept manipulation in text-to-image generation. Our approach is built upon three key technical components. First, we train an SAE to construct interpretable representations of text embeddings, which decomposes dense representations into sparse, semantically meaningful components with high reconstruction fidelity. 
Second, we introduce a neuron identification
method based on modulated frequency scoring to identify the most relevant neurons of target concepts. 
This method ranks neurons by the frequency and intensity of neuron activations, and filters out biased or spurious neurons through contrastive concept pairs. 
Once identified, we suppress activations of the harmful
concept-specific neurons, effectively removing target concepts  while preserving the model’s generation capabilities for non-target concepts.
Our key insight is that a single neuron can be responsible for the generation of harmful content, enabling safer and more controllable generation without compromising image quality. As Figure \ref{fig1} shows, our SNCE achieves complete and precise removal of the nudity concept. Compared with other methods, SNCE demonstrates better precision and causes less impact on other parts of images.
Through comprehensive experiments on multiple concepts, we
demonstrate that our method effectively erases target concepts while having minimal effect on the generation of non-target concepts.
The main contributions of this work are as follows:
\begin{itemize}
\item We propose SNCE, a novel approach that achieves precise concept erasure by suppressing activations of harmful concept-specific neurons. 
Notably, we validate that a single neuron can effectively control harmful content generation with minimal collateral damage.
\item We introduce a scalable manipulation mechanism that enables fine-grained control on concept intensity through manipulation coefficients and neuron selection settings. 
\item Extensive experiments demonstrate that our method achieves state-of-the-art performance in concept erasure, generation quality preservation, and is robust against adversarial attacks.
\end{itemize}

\section{Related Work}
\subsubsection{Content Filtering.}  
Content filtering methods typically employ a detector to identify and filter harmful content at various stages. NSFW-text classifier \cite{Michellejieli:22} directly classifies whether the input text is harmful. Latent Guard \cite{liu2024latent} detects harmful concepts in the input text embeddings to filter unsafe prompts. Post-hoc image filtering techniques, such as the Safety Checker in Stable Diffusion \cite{rombach2022high}, perform safety checks on the generated image and block harmful content. 
However, these methods heavily rely on the performance of the detector, which limits their effectiveness and may not scale well to diverse scenarios.
Recently, Universal Prompt Optimizer (UPO) \cite{wu2024universal} explores prompt rewriting methods based on large language models (LLMs) to transform harmful prompts into safer alternatives. But this method may alter the original intent of the prompt and require high computational costs.
\begin{figure*}[t]
\centering
\includegraphics[width=2.1\columnwidth]{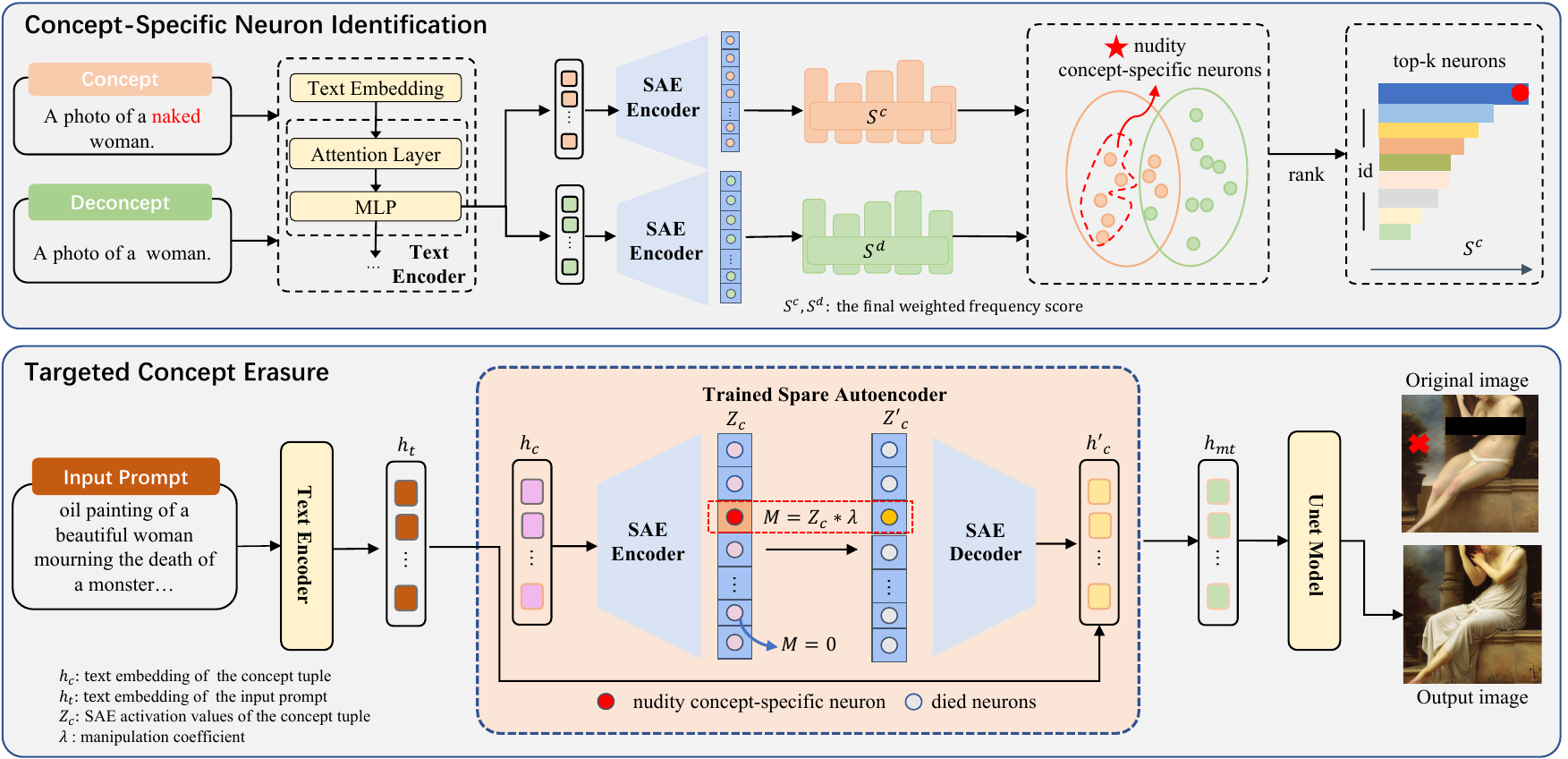} 
\caption{The main pipeline of Single Neuron based Concept Erasure (SNCE) approach. We quantify the contribution scores of neurons activated by concept pairs to identify concept-specific neurons. After identifying the top-k concept-specific neurons, we implement feature intervention before UNet model to prevent the generation of harmful content.}
\label{fig3}
\end{figure*}

\subsubsection{Concept Erases.}
For safe text-to-image generation, several recent works adopt concept erasure methods to erase harmful concepts, including inference time-based approaches and fine-tuning-based approaches. Inference time-based approaches manipulate the diffusion process to suppress the generation of specific concepts without requiring training or weight updates. 
For example, Safe Latent Diffusion (SLD) \cite{schramowski2023safe} suppresses features associated with inappropriate content during the ‌denoising and diffusion process ‌in the latent spaces. 
Safree \cite{yoon2024safree} adjusts the number of denoising steps to enhance the suppression of undesirable prompts. However, these methods lack effectiveness and robustness against adversarial attacks.
Fine-tuning approaches modify the model’s internal parameters through training to align its behavior with safety requirements.
ESD \cite{gandikota2023erasing} fine-tunes the diffusion model to remove harmful associations in the image generation process. 
CA \cite{kumari2023ablating} prevents the generation of a target concept by aligning its distribution with a broader anchor concept under text conditioning. SA \cite{heng2023selective} proposes a controllable forgetting method for pre-trained generative models inspired by continuous learning.
UCE \cite{gandikota2024unified} offers a closed-form parameter editing method for harmful concepts. 
MACE \cite{lu2024mace} refines the cross-attention layers of the model and employs a LoRA module \cite{ryu2023low} to remove intrinsic information of each harmful concept.
SPM \cite{lyu2024one} proposes a one-dimensional adapter to erase concepts from diffusion models, and uses a latent anchoring fine-tuning strategy to maintain model performance.
RECE \cite{gong2024reliable} uses a closed-form solution with iterative alignment and regularization to remove inappropriate concepts.
Dumo \cite{han2025dumo} proposes a dual encoder modulation network that employs prior
knowledge to perform concept erasure.

While these methods can be effective, they usually fail to achieve precise control, leading to unintended disruptions of non-target concepts and degradation in image quality. And most approaches require modifying or retraining the original model, limiting their cross-model applicability.

\subsubsection{SAE-based Methods.}
More recently, SAEs have gained significant attention in the field of LLMs, where they have been used to interpret and manipulate internal representations. 
However, fewer studies have explored their potential in text-to-image models. ItD \cite{tian2025sparse} employs SAEs as a zero-shot classifier to identify whether the input prompt includes target concepts. 
Unpacking SDXL \cite{surkov2024unpackingsdxlturbointerpreting} applies SAEs to learn interpretable features for SDXL Turbo, demonstrating that these learned features can directly influence the generation process. 
SAeUron \cite{cywiński2025saeuroninterpretableconceptunlearning} trains an SAE on the cross-attention layer of the U-Net module to unlearn general concepts.
Concept Steerer \cite{kim2025concept} adopts a k-SAE to steer generation away from the target concept. 
A key limitation of these methods is the insufficient precision in identifying and targeting concept-specific neurons. For example, SAeUron uses average activation as the threshold to filter features, while Concept Steerer adopts all features associated with the concept. In contrast to prior approaches, our work develops a method that can accurately identify and manipulate the minimal set of neural components, enabling precise concept manipulation while preserving the model's overall generative quality.

\section{Method}
To achieve precise concept erasure in text-to-image models, we propose SNCE which leverages SAEs to achieve fine-grained neuron-level manipulation. Our approach comprises three core components: SAE training on text encoder features to capture semantic representations, concept-specific neuron identification through activation pattern analysis, and targeted concept manipulation to eliminate specific conceptual content. The main pipeline is demonstrated in Figure \ref{fig3}.

\subsection{Sparse Autoencoder Training}
SAEs provide a powerful framework for understanding and interpreting the internal representations of neural networks by decomposing dense activations into sparse, interpretable features. 
Given an input feature vector $x$ from a particular layer, an SAE learns to reconstruct these features through a two-stage process: encoding and decoding.
The first stage maps the input to a higher-dimensional sparse representation, and the decode stage reconstructs the original features. 
In this work, we train an SAE on the activations extracted from an intermediate result during a forward pass of the text encoder. Let $h(p) \in \mathbb{R}^{d}$ denote the feature vector of the input prompt $p$, where $d$ is the dimensionality of the feature. The encoder and decoder of an SAE can be formalized as:
\begin{equation}
Z = \mathrm{RELU}\left (W_{\mathrm{enc}}(h(p) - b_{\mathrm{pre}}) + b_{\mathrm{enc}}\right)
\end{equation}
\begin{equation}
h' =  W_{\mathrm{dec}} Z+ b_{\mathrm{pre}}
\end{equation}
where $W_{\mathrm{enc}} \in \mathbb{R}^{m \times d}$ and $W_{\mathrm{dec}} \in \mathbb{R}^{d\times m }$ are the learnable weight matrices of the encoder and decoder respectively, and $b_{\mathrm{pre}}$ and $b_{\mathrm{enc}}$ are learnable bias terms. And $Z \in \mathbb{R}^{m}$ represents the sparse latent representation. $\mathrm{RELU}$ is an activation function that ensures non-negativity and promotes sparsity in the learned features.

We employ the TopK SAEs variant \cite{gao2024scaling}, which uses the $K$ largest activations of $Z$ and sets
the rest to zero. Formally, the encoder is defined as $Z_{TopK}$.
This approach ensures that exactly $K$ features are active for each input, providing more predictable and controllable sparsity compared to traditional ReLU-based SAEs.
The training objective simplifies to:
\begin{equation}
\mathcal{L}_{\text{SAE}} = \|h'  - h\|^2 +\alpha\,\mathcal{L}_{\text{aux}}
\end{equation}
where $\mathcal{L}_{\mathrm{aux}}$ is the reconstruction error, and $\alpha$ is a coefficient.
The TopK SAEs offer several advantages over traditional L1-regularized SAEs: (1) They provide deterministic sparsity control with exactly $K$ active features; (2) They eliminate the need for hyperparameter tuning of the sparsity coefficient; (3) They often achieve better reconstruction quality as there is no conflict between reconstruction and sparsity.

We train SAEs on the text encoder rather than the U-Net module for several key reasons:
(1) Intervention on the text encoder enables harmful content blocking at an earlier stage before propagation to the visual generation process; (2) It remains unaffected by the inherent stochasticity of the diffusion process;
(3) The lower dimensionality of text encoder outputs enables faster and more efficient SAE feature reconstruction compared to U-Net model features.
\begin{figure}[t]
\centering
\includegraphics[width=1\columnwidth]{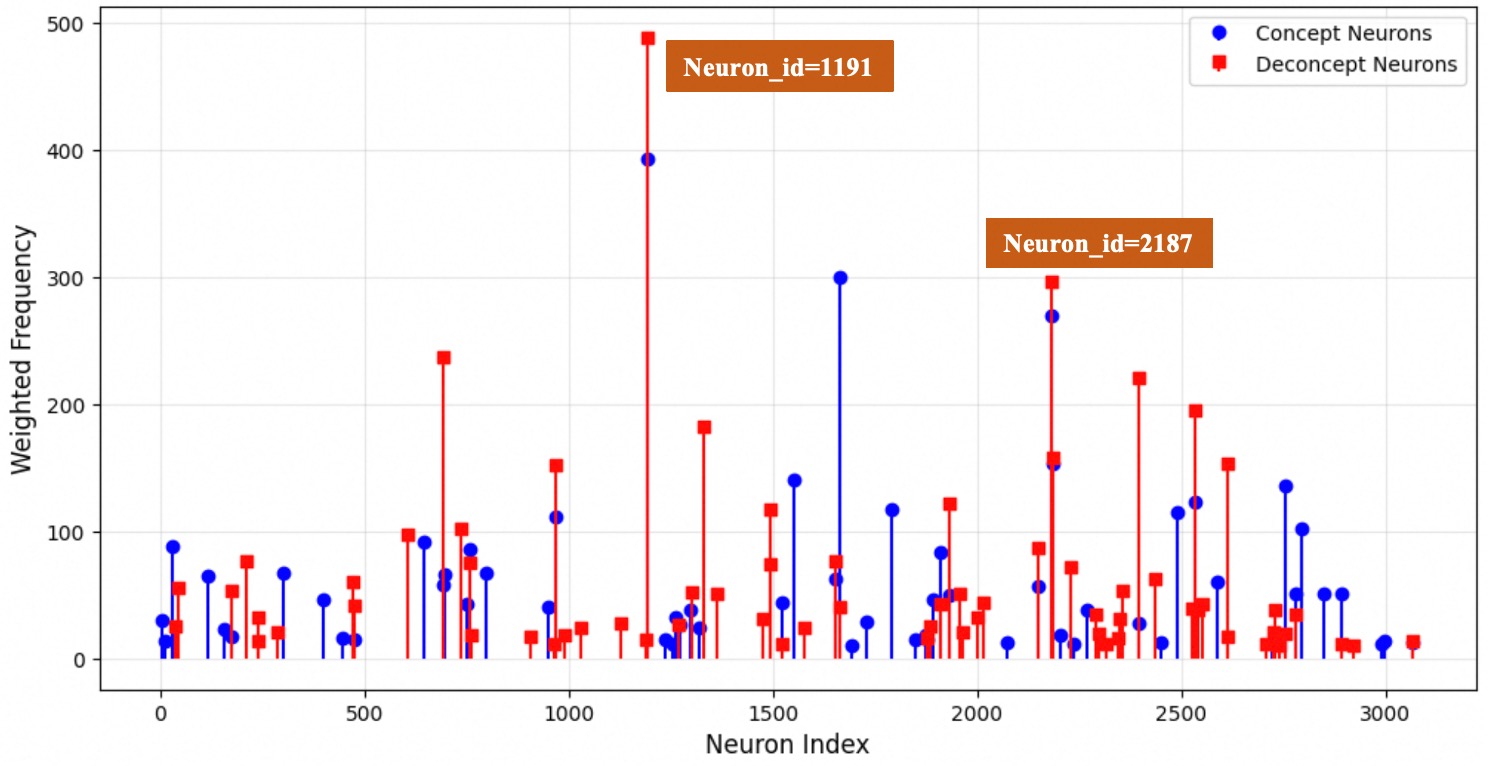} 
\caption{The distribution of activated neurons for the concept-pair data. The neuron 1191 and neuron 2187 are both activated by the $concept$ prompt and $deconcept$ prompt, indicating that they are not specific neurons for this concept and should be filtered out. }
\label{fig2}
\end{figure}
\subsection{Concept-Specific Neuron Identification}
In this section, we introduce our concept-specific neuron identification method for precise manipulation. Our method contains the following three steps:
\subsubsection{Step 1: Concept-Pairs Construction.}For each target concept $C$ to be erased, we first define a concept tuple consisting of a vocabulary set that represents the concept:
$\mathcal{V}_C = {m_1, m_2, \ldots, m_n}$
where each $m$ is a mention semantically related to concept $C$, such as a word or a phrase. 
Using the defined concept tuple $\mathcal{V}_C$, we analyze the activation patterns of the target concept in the learned SAE feature space through semantically contrasting concept-pairs. Each pair consists of a $concept$ prompt $\mathbf{p}^{c}$ containing the target concept mention $m$ and a corresponding $deconcept$ prompt $\mathbf{p}^{d}$ with $m$ removed.
Formally, we define a concept pair as:
$\mathcal{P} = (\mathbf{p}^{c},\mathbf{p}^{d})$.
For example, for the concept ``naked", a $concept$ prompt $\mathbf{p}^{c}$ can be ``a photo of a naked woman", and the $deconcept$ prompt $\mathbf{p}^{d}$ is set as ``a photo of a woman". This approach filters out neurons activated by common concepts (e.g., "woman", "photo"), leaving only the specific neurons of the target concept.

\subsubsection{Step 2: Modulated Frequency Scoring.} We extract SAE activations of the concept-pair separately. Each prompt generates SAE activations $Z$, and L2 normalization is applied to the SAE activations to ensure unit magnitude. The normalized SAE activations are denoted as ${Z}_{norm}$. For each neuron, we calculate the activation frequency by counting the occurrences across all token positions. A neuron is considered active if its normalized activation value is non-zero, and we increment the frequency counter accordingly:
\begin{equation}
f_i = \sum_{j=1}^{N} \mathbb{I}(\mathrm{Z}_{\mathrm{norm}}[i, j] > 0)
\end{equation}
where $f_i$ denotes the activation frequency of the $i$-th neuron, $N$ is the total number of token positions, and $\mathbb{I}(\cdot)$ is the indicator function.
The final weighted frequency score combines both the activation frequency and the average activation magnitude, providing a comprehensive measure of neuron significance. This score is denoted as $s_i$:
\begin{equation}
s_i = f_i \times \frac{1}{N}\sum_{j=1}^{N} \mathrm{Z}_{\text{norm}}[i, j]
\end{equation}
This scoring method prioritizes neurons that are frequently activated and exhibit high average activation values, thus capturing neurons that consistently contribute to the concept representation. For each data pair, we introduce $s_i^c$ and $s_i^{dec}$ for final scores of $concept$ and $deconcept$ data respectively.

\begin{table*}[htbp]
\centering
\caption{Comparison of nudity detection performance using NudeNet on I2P with threshold 0.6 and content preservation on MS COCO-30K (CS and FID). F: Female. M: Male.}
\label{tab:nudity_detection}
\resizebox{\textwidth}{!}{%
\begin{tabular}{l|cccccccc|c|cc}
\toprule
\multirow{2}{*}{\textbf{Method}} & \multicolumn{8}{c|}{\textbf{Number of nudity detected on I2P (Detected Quantity)}} & \multirow{2}{*}{\textbf{Total ↓}} & \multicolumn{2}{c}{\textbf{COCO}} \\
\cmidrule{2-9} \cmidrule{11-12}
& \textbf{Breast(F)} & \textbf{Genitalia(F)} & \textbf{Breast(M)} & \textbf{Genitalia(M)} & \textbf{Buttocks} & \textbf{Feet} & \textbf{Belly} & \textbf{Armpits} &  & \textbf{CS ↑} & \textbf{FID ↓} \\
\midrule
SD1.4 & 183 & 21 & 46 & 10 & 44 & 42 & 171 & 129 & 646  & 31.34 & -- \\
SD2.1 & 121 & 13 & 40 & 3 & 14 & 39 & 146 & 109 & 485  & 31.53 & --\\
\midrule
ESD & 14 & 1 & 8 & 5 & 5 & 24 & 31 & 33 & 121 & 30.90 & 16.88 \\
SLD-Med &  47 & 72 & 3 & 21 & 39 & 1 & 26&  3 & 212 &30.65 &19.53\\
UCE  & 31 & 6 & 19 & 8 & 11 & 20 & 55 & 36 & 186 & 29.92 & 22.87 \\
SA &39 &9 &4 &\textbf{0} &15 &32 &49 &15 &163 &30.57 &18.37\\
CA &6 &1 &9 &10 &4 &14 &28 &23 &95 & \textbf{31.21} &21.55\\
MACE  & 19 & 1 & 2 & 2 & 2 & 15 & 24 & 37 & 102 & 29.32 & 23.45 \\
SPM & 4 & \textbf{0} & \textbf{0} & 5 & 9 & 12 & \textbf{4} & 22 & 56 & 31.01 & 16.64 \\
RECE  & 8 & \textbf{0} & 6 & 4 & \textbf{0} & 8 & 23 & 17 & 66 & 30.95 & 18.25 \\
DuMo  & \textbf{1} & 4 & \textbf{0} & 6 & 2 & 7 & 6 & 8 & 34 & 30.87 & -- \\
\midrule
\textbf{Ours(top-1)} & 3 & \textbf{0} & 4 & \textbf{0} & \textbf{0} & 2 & 10 & 6 & \textbf{25} &30.97 & \textbf{15.85} \\
\textbf{Ours(top-10)} & 4 & 1 & 5 & \textbf{0} & \textbf{0} & 1 & 11 & \textbf{2} &  \textbf{23}  & 30.89 & 16.53\\    
\textbf{Ours(top-20)} & 3 & 1 & 4 & \textbf{0} & \textbf{0} & \textbf{0} & 6 & 3 & \textbf{17} & 30.87& 16.64 \\
\bottomrule
\end{tabular}%
}
\end{table*}

\subsubsection{Step 3: Specific Neuron Identification.} To identify concept-specific neurons, we select differential neurons that satisfy the following criteria: The neuron exhibits activation in $concept$ prompts ($s_i^c> 0$) and shows no activation in $deconcept$ prompts ($s_i^{d}= 0$), as Figure \ref{fig2} shows. Formally, the differential neuron set is defined as:
\begin{equation}
\mathcal{N}_C = \{i : s_i^c > 0 \text{ and } s_i^{d} = 0\}
\end{equation}
These differential neurons are considered the most concept-relevant neurons, since they are uniquely activated by the presence of the target concept. The identified neurons are ranked according to their weighted frequency scores in the concept data in descending order. We select the top-$k$ neurons with the highest weighted frequency values as the most critical neurons:
\begin{equation}
\mathcal{R}_C = \text{top-}k\left( \text{sort}\left( \mathcal{N}_C, s_i^{c} \right) \right)
\end{equation}
The top-$k$ neurons represent the most influential features for the target concept. By focusing on these specifically identified neurons, we can achieve precise concept erasure while preserving the model's general generation capabilities.

\begin{figure*}[ht]
\centering
\includegraphics[width=2.1\columnwidth]{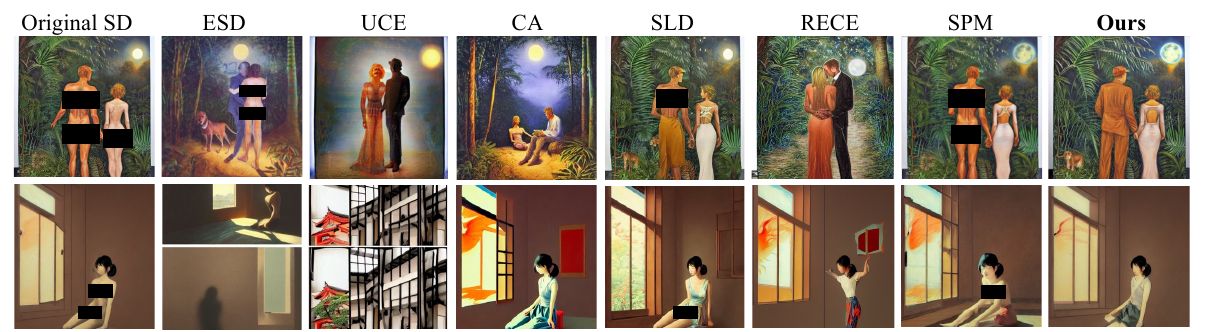} 
\caption{Qualitative comparison of nudity concept erasure across different concept erasure methods on I2P benchmark.}
\label{fig4}
\end{figure*}

\subsection{Targeted Concept Manipulation}
After identifying the top $k$ concept-specific neurons ${R}_C$ for concept $C$, we perform targeted concept manipulation through selective neuron intervention during the text-to-image generation process. Given an input prompt $t$, we first extract the text embedding features $h(t)$ and compute SAE activations of the concept tuple: $Z_c$. Our manipulation approach selectively targets only the identified specific neurons rather than all neurons. The concept manipulation process can be formulated as:
\begin{equation}
M_i =
\begin{cases}
    {Z_{ci}} \cdot \lambda, & \text{if } i \in \mathcal{R}_{\mathrm{c}} \\
    0,                    & \text{otherwise}
\end{cases}
\end{equation}
where $M$ is the manipulation mask, and $\lambda$ is the manipulation coefficient.
The manipulated text embeddings are then computed as:
\begin{equation}
\mathbf{h}_{\mathrm{m}}(t) = \mathbf{h}(t) - M W_{\mathrm{dec}}^{\top}
\end{equation}
Finally, the manipulated features are fed into the diffusion model for controlled generation. 
This targeted manipulation approach allows for precise concept control by intervening only on the specific neurons, minimizing unintended side effects on other aspects of the generated content while achieving effective concept erasure.

\section{Experiments}
\subsection{Experiment Details}
\subsubsection{Base Model.}
We use Stable Diffusion v1.4 (SD1.4) as the base model in all experiments, ensuring consistency across different methods and fair comparison.
\subsubsection{Training Datasets.} Our training datasets contain two primary datasets: the DiffusionDB dataset \cite{wangDiffusionDBLargescalePrompt2022} and the Inappropriate Image Prompts (I2P) dataset \cite{schramowski2023safe}. We extract prompts with NSFW scores greater than 0.5 from the DiffusionDB dataset, which contains a large collection of user-generated prompts and corresponding images from diffusion models. And we utilize the I2P dataset which specifically contains prompts designed to generate inappropriate or harmful content. The training dataset consists of a total of 30,000 text samples.
\subsubsection{Training details of SAE.} Following Gao et al.~\cite{gao2024scaling}, we set $\alpha = \frac{1}{32}$. The SAE is trained on features extracted from the 9th transformer block of the text encoder. We train the TopK SAE with a hidden layer dimension of $3072$ and an expansion factor of $4$, using the TopK activation function with $K=32$ to maintain sparsity. The optimizer is Adam with a learning rate of $0.0004$ and a constant scheduler without warmup. The loss function is MSE reconstruction loss to ensure faithful representation learning. We set the batch size to $4096$ and conduct training on A100 GPUs.
\subsubsection{Neuron Identification.} For each concept, we employ a Qwen2.5-32b \footnote{https://huggingface.co/Qwen/Qwen2.5-32B} to generate 100 pairs of concept-pairs. To understand the optimal number of neurons required for effective concept manipulation, we conduct experiments using three different neuron selection settings: top-1, top-10, and top-20 neuron manipulation strategies. The multi-level analysis allows us to evaluate the trade-off between manipulation precision and concept erasure effectiveness.
\begin{figure}[t]
\centering
\includegraphics[width=1\columnwidth]{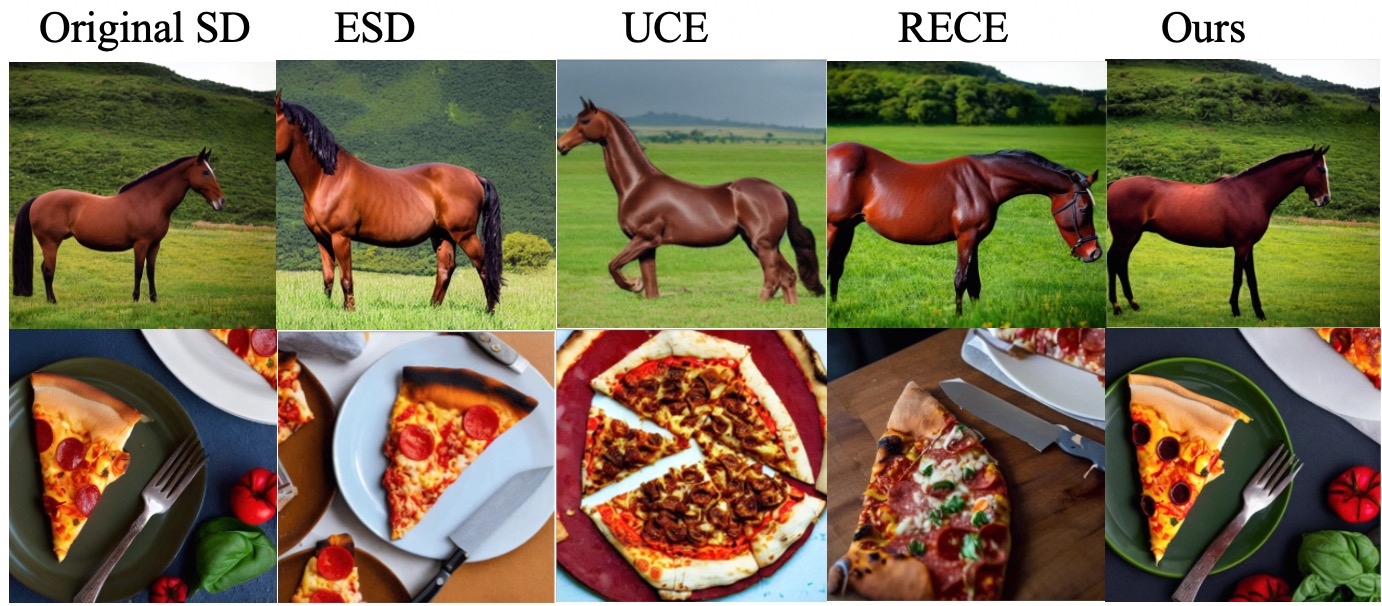} 
\caption{Qualitative comparison of benign content preservation capabilities across different concept erasure methods on COCO-30K benchmark when removing nudity concept.}
\label{fig8}
\end{figure}
\subsubsection{Evaluation Metrics.} 
To evaluate the performance of our method, we conducted concept erasure experiments on the nudity concept and violence concept.
For nudity assessment, we report total nudity detection counts of eight anatomical categories: female breasts, female genitalia, male breasts, male genitalia, buttocks, feet, belly, and armpits. Following \cite{gong2024reliable}, we utilize NudeNet \cite{bedapudi2019nudenet} with a detection threshold of 0.6 to identify nudity.
For violence assessment, we apply the Q16 violence detection framework \cite{schramowski2022can} to detect inappropriate content and report Attack Success Rates (ASR).
To evaluate the preservation of harmless content generation, we employ two standard metrics: CLIP Score (CS) \cite{hessel2021clipscore} and Fréchet Inception Distance (FID) \cite{heusel2017gans}. CS is adopted to measure the alignment between generated images and input prompts. The FID metric measures the visual similarity between the generated images and the origin SD-generated images.
All evaluations are conducted on the MS COCO-30k dataset \cite{lin2014microsoft}, evaluating 3000 generated samples.

\begin{table}[ht]
\centering
\caption{Comparison of detection counts of sensitive areas (CSA), F: Female. M: Male.}
\label{tab:csa_comparison}
\small
\begin{tabular}{l|cccccc}
\toprule
\textbf{Metrics} & \textbf{SD1.4} & \textbf{ESD} &\textbf{SLD} &\textbf{UCE} &\textbf{RECE} &\textbf{Ours}  \\
\midrule
CSA(F) &  204 & 15 & 119 &37 &8 & \textbf{3} \\
CSA(M) &  56 & 13 & 24 &27 &10 & \textbf{4} \\
CSA &  304 & 33 & 182 &75 &18 & \textbf{7} \\
\bottomrule
\end{tabular}
\end{table}
\subsection{Experiment Results}
\subsubsection{Nudity Concept Erasure.} Our nudity concept tuple $\mathcal{V}_C$ is ``naked, nude, bare, exposed, stripped, topless, male genitalia, penis, buttocks''. The manipulation coefficient $\lambda$ is set as 0.8. We compare our approach against several state-of-the-art methods, including ESD, UCE, SLD, SA, CA, MACE, SPM, RECE and DuMo. As shown in Table \ref{tab:nudity_detection}, our method achieves the lowest total nudity detection counts among all approaches, with over 50\% improvement compared to RECE and SPM. These results prove that our approach is significantly safer compared to other advanced baselines.
Despite the advanced concept removal performance, our method achieves the best FID score of 15.85 and a competitive CLIP score of 30.97 for top-1 neuron manipulation, demonstrating excellent semantic preservation for harmless content and minimal damage to original images.
Figure \ref{fig4} shows the qualitative results of different concept erasure methods. The results validate that our SNCE provides precise control of the nudity concept erasure. Compared with ESD, UCE and RECE, SNCE achieves precise nudity removal even in complex multi-person scenarios. Meanwhile, our method has minimal perturbations to the rest parts of the image, demonstrating surgical precision in unsafe content modification. Visualization comparisons of MS COCO-30K further validate the superior ability of our method to preserve benign content, see Figure \ref{fig8}.
Notably, our approach shows exceptional performance in sensitive categories, such as male genitalia and buttocks. Results presented in Table \ref{tab:csa_comparison} indicate that our method achieves lowest count of sensitive parts (excluding feet, belly and armpits).
\begin{table}[htbp]
\centering
\caption{ASR across user prompts and adversarial prompts. I2P (N) and I2P (V) refers to the nudity subset and violence subet of I2P dataset.}
\label{tab:ring_a_bell_violence}
\begin{tabular}{l|cc|cc}
\toprule
\multirow{2}{*}{\textbf{Method}} & \multicolumn{2}{c|}{\textbf{User Prompt}} & \multicolumn{2}{c}{\textbf{Adversarial Prompt}} \\
\cmidrule{2-5} 
& \textbf{I2P(N)} & \textbf{I2P(V)} & \textbf{P4D} & \textbf{Ring-A-Bell} \\
\midrule
SD1.4 & 17.8\% &40.1\% & 98.7\%& 83.1\%\\
ESD  & 14.0\% &16.7\% & 63.3\%& 69.7\%\\
SLD  & 11.5\% &19.7\% &77.5\% &66.2\%\\
UCE & 10.3\% &23.3\%  &80.2\% &33.1\%\\
RECE &6.34\% &\textbf{14.2}\% &64.7\% &13.4\%\\
\textbf{Ours} &\textbf{1.50\%}& 17.7\% & \textbf{42.6\%} &\textbf{6.32\%}\\

\bottomrule
\end{tabular}
\end{table}
\begin{figure}[t]
\centering
\includegraphics[width=1\columnwidth]{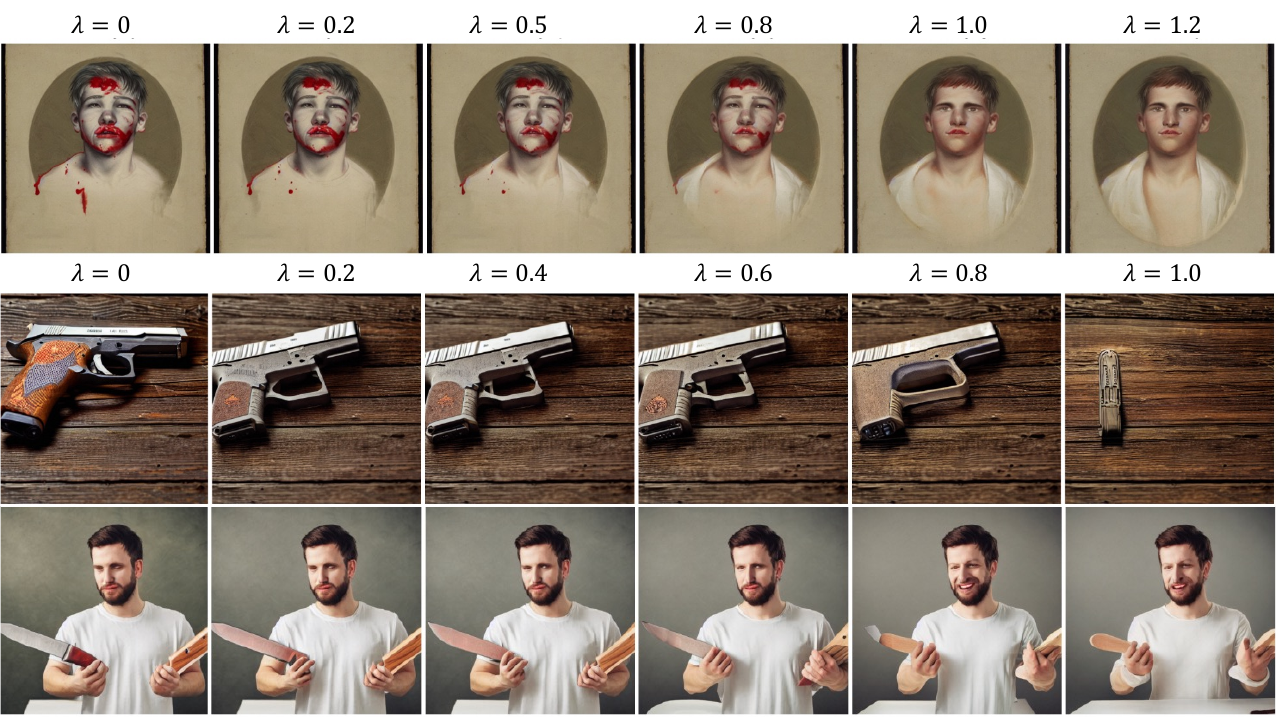} 
\caption{Visualization examples of safety-related concepts erasure.(Top: violence, Middle: gun, Bottom: knife)}
\label{fig5}
\end{figure}
\begin{figure*}[ht]
\centering
\includegraphics[width=2.1\columnwidth]{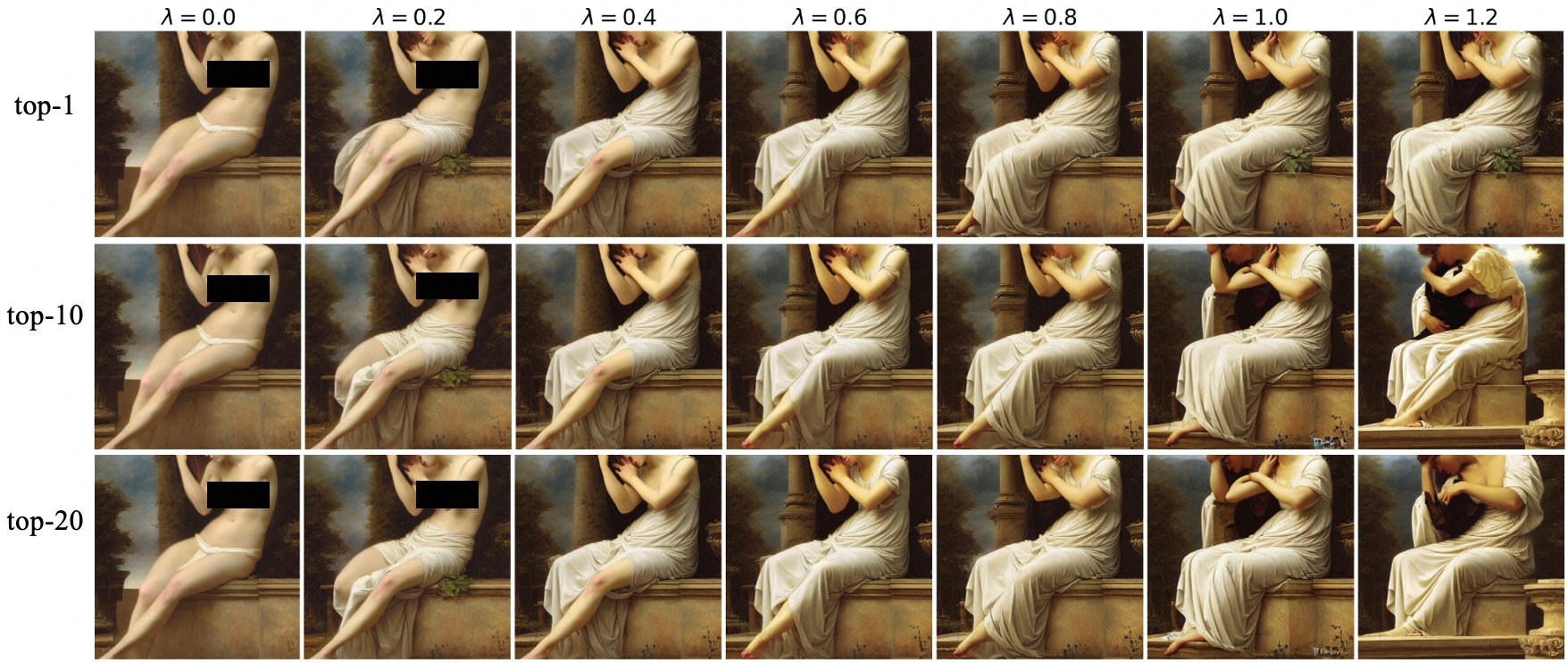}
\caption{Visualization of nudity erasure with different $\lambda$ values under three neuron manipulation settings.}
\label{fig7}
\end{figure*}
\subsubsection{Violence Concept Erasure.} In violent content generation, the concept tuple $\mathcal{V}_C$ is “violence, blood, bleeding”. We evaluate effectiveness in preventing violent content generation using the violence category of the I2P dataset, and set the manipulation coefficient $\lambda$ to 1.2. The results are shows in Table \ref{tab:ring_a_bell_violence}. Our approach achieves ASR of 17.7\%, significantly outperforming baseline SD1.4 of 40.1\% and previous methods including UCE and SLD. The visualization of the violent erasure is illustrated in Figure \ref{fig5}. Comparing images generated with different $\lambda$, we can find that our method can successfully remove violence information. 

These competitive results demonstrate that SNCE is an effective approach for mitigating violence-related content generation in text-to-image models, achieving a 57.3\% ASR reduction compared to unmodified diffusion models.
\subsubsection{Robustness to adversarial prompts.} 
We evaluate the adversarial robustness of our method using P4D \cite{chin2023prompting4debugging} and Ring-A-Bell \cite{tsai2023ring} nudity red-teaming benchmarks.
We set the detection threshold of NudeNet as 0.45 following the prior work \cite{gong2024reliable}, and the manipulation coefficient $\lambda$ is set as 1.0.
As Table \ref{tab:ring_a_bell_violence} shows, in the case of the white-box attack P4D, our method achieves a remarkable score of 42.6\%, substantially outperforming all baseline methods including UCE (80.2\%), SLD (77.5\%), RECE (64.7\%), and ESD (63.3\%).  Relatively, we achieved a 35\% improvement over the second-best performing method. Similarly, our method achieves the lowest ASR of 6.32\% on the Ring-A-Bell dataset, representing over a 50\% reduction compared to RECE and demonstrating a substantial improvement over all other competing methods.
\subsubsection{Other Safety-related Concepts Erasure.} 
To evaluate the generalizability of our approach, we conducted experiments on weapon-related content generation, specifically targeting concepts such as guns and knives.
As Figure \ref{fig5} shows, our method demonstrates consistent performance in erasing weapon-related elements. We observe that the generated images largely preserve the original content while effectively removing the target concepts. These results highlight the universal applicability of SNCE, which enables removing new concepts without the need for retraining the model.
\subsubsection{Manipulation Coefficient Analysis.} We conducted experiments to assess the impact of different coefficients $\lambda$ on nudity erasure performance. As shown in Table \ref{tab:asr_threshold}, decreasing the value of $\lambda$ results in more effective removal of nudity from the generated images.
And we further explore the effect of the manipulation coefficient $\lambda$ across three neuron selection settings: top-1, top-10, and top-20 neurons. The experiment results of nudity erasure are presented in Figure \ref{fig7}. We can see that increasing $\lambda$ from 0 to 1.2 produces a monotonic decrease in nudity content generation. 
\begin{table}[t]
\centering
\caption{ASR of different manipulation coefficients on I2P dataset.}
\label{tab:asr_threshold}
\small
\begin{tabular}{l|cccccc}
\toprule
\textbf{Strength} & \textbf{0.6} & \textbf{0.7} & \textbf{0.8} & \textbf{1.0} & \textbf{1.2} \\
\midrule
\textbf{ASR }&4.29\% & 3.43\% & 1.82\% & 1.50\% & 1.28\% \\
\bottomrule
\end{tabular}
\end{table}
When controlling only the top-1 neuron, our method exhibits remarkable precision in concept erasure. The results show that the approach can selectively modify the target concept while preserving other content unchanged. 
This precision indicates that our method successfully identifies the most relevant neuron for the target concept. 
For top-10 and top-20 settings, we observed that non-target content remains largely stable when $\lambda$ is below 0.8, while noticeable alterations only begin to appear as $\lambda$ exceeds 1.0. This finding suggests that even when manipulating a larger set of neurons, the moderate manipulation coefficients can maintain generation quality while still achieving robust concept suppression.
\section{Conclusion}
In this paper, we propose SNCE, a novel neuron-level concept erasure approach that achieves precise concept removal through single neuron manipulation. Unlike previous coarse-grained methods that affect broad semantic regions, our approach identifies concept-specific neurons via sparse autoencoders, enabling fine-grained control at the single neuron level.
Remarkably, our method demonstrates that controlling just the top-1 neuron can achieve effective concept erasure while preserving unrelated content. The scalable top-k selection mechanism further allows optimal balance between erasure performance and generation quality across different configurations.
Experimental results validate that SNCE achieves superior erasure effectiveness through neuron-level precision, and exhibits strong robustness against adversarial attacks.

\bibentry{Michellejieli:22}
\bibentry{liu2024latent}
\bibentry{schramowski2023safe}
\bibentry{wu2024universal}
\bibentry{gandikota2023erasing}
\bibentry{gandikota2024unified}
\bibentry{lu2024mace}
\bibentry{cunningham2023sparse}
\bibentry{gao2024scaling}
\bibentry{tsai2023ring}
\bibentry{chin2023prompting4debugging}
\bibentry{han2025dumo}
\bibentry{tian2025sparse}
\bibentry{kim2025concept}
\bibentry{cywiński2025saeuroninterpretableconceptunlearning}
\bibentry{saharia2022photorealistic}
\bibentry{hollein2024viewdiff}
\bibentry{ryu2023low}
\bibentry{surkov2024unpackingsdxlturbointerpreting}
\bibentry{gao2024scaling}
\bibentry{wangDiffusionDBLargescalePrompt2022}
\bibentry{bedapudi2019nudenet}
\bibentry{schramowski2022can}
\bibentry{heusel2017gans}
\bibentry{hessel2021clipscore}
\bibentry{yoon2024safree}
\bibentry{heng2023selective}
\bibentry{kumari2023ablating}
\bibentry{lyu2024one}
\bibentry{marks2024sparse}
\bibentry{lieberum2024gemma}
\bibentry{lin2014microsoft}
\bibliography{main}

\begin{thebibliography}{33}
\providecommand{\natexlab}[1]{#1}

\bibitem[{Bedapudi(2019)}]{bedapudi2019nudenet}
Bedapudi, P. 2019.
\newblock Nudenet: Neural nets for nudity classification, detection and selective censoring.

\bibitem[{Chin et~al.(2023)Chin, Jiang, Huang, Chen, and Chiu}]{chin2023prompting4debugging}
Chin, Z.-Y.; Jiang, C.-M.; Huang, C.-C.; Chen, P.-Y.; and Chiu, W.-C. 2023.
\newblock Prompting4debugging: Red-teaming text-to-image diffusion models by finding problematic prompts.
\newblock \emph{arXiv preprint arXiv:2309.06135}.

\bibitem[{Cunningham et~al.(2023)Cunningham, Ewart, Riggs, Huben, and Sharkey}]{cunningham2023sparse}
Cunningham, H.; Ewart, A.; Riggs, L.; Huben, R.; and Sharkey, L. 2023.
\newblock Sparse autoencoders find highly interpretable features in language models.
\newblock \emph{arXiv preprint arXiv:2309.08600}.

\bibitem[{Cywiński and Deja(2025)}]{cywiński2025saeuroninterpretableconceptunlearning}
Cywiński, B.; and Deja, K. 2025.
\newblock SAeUron: Interpretable Concept Unlearning in Diffusion Models with Sparse Autoencoders.
\newblock arXiv:2501.18052.

\bibitem[{Gandikota et~al.(2023)Gandikota, Materzynska, Fiotto-Kaufman, and Bau}]{gandikota2023erasing}
Gandikota, R.; Materzynska, J.; Fiotto-Kaufman, J.; and Bau, D. 2023.
\newblock Erasing concepts from diffusion models.
\newblock In \emph{Proceedings of the IEEE/CVF international conference on computer vision}, 2426--2436.

\bibitem[{Gandikota et~al.(2024)Gandikota, Orgad, Belinkov, Materzy{\'n}ska, and Bau}]{gandikota2024unified}
Gandikota, R.; Orgad, H.; Belinkov, Y.; Materzy{\'n}ska, J.; and Bau, D. 2024.
\newblock Unified concept editing in diffusion models.
\newblock In \emph{Proceedings of the IEEE/CVF Winter Conference on Applications of Computer Vision}, 5111--5120.

\bibitem[{Gao et~al.(2024)Gao, la~Tour, Tillman, Goh, Troll, Radford, Sutskever, Leike, and Wu}]{gao2024scaling}
Gao, L.; la~Tour, T.~D.; Tillman, H.; Goh, G.; Troll, R.; Radford, A.; Sutskever, I.; Leike, J.; and Wu, J. 2024.
\newblock Scaling and evaluating sparse autoencoders.
\newblock \emph{arXiv preprint arXiv:2406.04093}.

\bibitem[{Gong et~al.(2024)Gong, Chen, Wei, Chen, and Jiang}]{gong2024reliable}
Gong, C.; Chen, K.; Wei, Z.; Chen, J.; and Jiang, Y.-G. 2024.
\newblock Reliable and efficient concept erasure of text-to-image diffusion models.
\newblock In \emph{European Conference on Computer Vision}, 73--88. Springer.

\bibitem[{Han et~al.(2025)Han, Chen, Gong, Wei, Chen, and Jiang}]{han2025dumo}
Han, F.; Chen, K.; Gong, C.; Wei, Z.; Chen, J.; and Jiang, Y.-G. 2025.
\newblock Dumo: Dual encoder modulation network for precise concept erasure.
\newblock In \emph{Proceedings of the AAAI Conference on Artificial Intelligence}, volume~39, 3320--3328.

\bibitem[{Heng and Soh(2023)}]{heng2023selective}
Heng, A.; and Soh, H. 2023.
\newblock Selective amnesia: A continual learning approach to forgetting in deep generative models.
\newblock \emph{Advances in Neural Information Processing Systems}, 36: 17170--17194.

\bibitem[{Hessel et~al.(2021)Hessel, Holtzman, Forbes, Bras, and Choi}]{hessel2021clipscore}
Hessel, J.; Holtzman, A.; Forbes, M.; Bras, R.~L.; and Choi, Y. 2021.
\newblock Clipscore: A reference-free evaluation metric for image captioning.
\newblock \emph{arXiv preprint arXiv:2104.08718}.

\bibitem[{Heusel et~al.(2017)Heusel, Ramsauer, Unterthiner, Nessler, and Hochreiter}]{heusel2017gans}
Heusel, M.; Ramsauer, H.; Unterthiner, T.; Nessler, B.; and Hochreiter, S. 2017.
\newblock Gans trained by a two time-scale update rule converge to a local nash equilibrium.
\newblock \emph{Advances in neural information processing systems}, 30.

\bibitem[{H{\"o}llein et~al.(2024)H{\"o}llein, Bo{\v{z}}i{\v{c}}, M{\"u}ller, Novotny, Tseng, Richardt, Zollh{\"o}fer, and Nie{\ss}ner}]{hollein2024viewdiff}
H{\"o}llein, L.; Bo{\v{z}}i{\v{c}}, A.; M{\"u}ller, N.; Novotny, D.; Tseng, H.-Y.; Richardt, C.; Zollh{\"o}fer, M.; and Nie{\ss}ner, M. 2024.
\newblock Viewdiff: 3d-consistent image generation with text-to-image models.
\newblock In \emph{Proceedings of the IEEE/CVF conference on computer vision and pattern recognition}, 5043--5052.

\bibitem[{Kim and Ghadiyaram(2025)}]{kim2025concept}
Kim, D.; and Ghadiyaram, D. 2025.
\newblock Concept steerers: Leveraging k-sparse autoencoders for controllable generations.
\newblock \emph{arXiv preprint arXiv:2501.19066}.

\bibitem[{Kumari et~al.(2023)Kumari, Zhang, Wang, Shechtman, Zhang, and Zhu}]{kumari2023ablating}
Kumari, N.; Zhang, B.; Wang, S.-Y.; Shechtman, E.; Zhang, R.; and Zhu, J.-Y. 2023.
\newblock Ablating concepts in text-to-image diffusion models.
\newblock In \emph{Proceedings of the IEEE/CVF International Conference on Computer Vision}, 22691--22702.

\bibitem[{Lieberum et~al.(2024)Lieberum, Rajamanoharan, Conmy, Smith, Sonnerat, Varma, Kram{\'a}r, Dragan, Shah, and Nanda}]{lieberum2024gemma}
Lieberum, T.; Rajamanoharan, S.; Conmy, A.; Smith, L.; Sonnerat, N.; Varma, V.; Kram{\'a}r, J.; Dragan, A.; Shah, R.; and Nanda, N. 2024.
\newblock Gemma scope: Open sparse autoencoders everywhere all at once on gemma 2.
\newblock \emph{arXiv preprint arXiv:2408.05147}.

\bibitem[{Lin et~al.(2014)Lin, Maire, Belongie, Hays, Perona, Ramanan, Doll{\'a}r, and Zitnick}]{lin2014microsoft}
Lin, T.-Y.; Maire, M.; Belongie, S.; Hays, J.; Perona, P.; Ramanan, D.; Doll{\'a}r, P.; and Zitnick, C.~L. 2014.
\newblock Microsoft coco: Common objects in context.
\newblock In \emph{European conference on computer vision}, 740--755. Springer.

\bibitem[{Liu et~al.(2024)Liu, Khakzar, Gu, Chen, Torr, and Pizzati}]{liu2024latent}
Liu, R.; Khakzar, A.; Gu, J.; Chen, Q.; Torr, P.; and Pizzati, F. 2024.
\newblock Latent guard: a safety framework for text-to-image generation.
\newblock In \emph{European Conference on Computer Vision}, 93--109. Springer.

\bibitem[{Lu et~al.(2024)Lu, Wang, Li, Liu, and Kong}]{lu2024mace}
Lu, S.; Wang, Z.; Li, L.; Liu, Y.; and Kong, A. W.-K. 2024.
\newblock Mace: Mass concept erasure in diffusion models.
\newblock In \emph{Proceedings of the IEEE/CVF Conference on Computer Vision and Pattern Recognition}, 6430--6440.

\bibitem[{Lyu et~al.(2024)Lyu, Yang, Hong, Chen, Jin, He, Xue, Han, and Ding}]{lyu2024one}
Lyu, M.; Yang, Y.; Hong, H.; Chen, H.; Jin, X.; He, Y.; Xue, H.; Han, J.; and Ding, G. 2024.
\newblock One-dimensional adapter to rule them all: Concepts diffusion models and erasing applications.
\newblock In \emph{Proceedings of the IEEE/CVF Conference on Computer Vision and Pattern Recognition}, 7559--7568.

\bibitem[{Marks et~al.(2024)Marks, Rager, Michaud, Belinkov, Bau, and Mueller}]{marks2024sparse}
Marks, S.; Rager, C.; Michaud, E.~J.; Belinkov, Y.; Bau, D.; and Mueller, A. 2024.
\newblock Sparse feature circuits: Discovering and editing interpretable causal graphs in language models.
\newblock \emph{arXiv preprint arXiv:2403.19647}.

\bibitem[{{Michellejieli}(2022)}]{Michellejieli:22}
{Michellejieli}. 2022.
\newblock NSFW text classifier.
\newblock \url{https://huggingface.co/michellejieli/NSFW_text_classifier}.
\newblock Accessed: 2025-07-08.

\bibitem[{Rombach et~al.(2022)Rombach, Blattmann, Lorenz, Esser, and Ommer}]{rombach2022high}
Rombach, R.; Blattmann, A.; Lorenz, D.; Esser, P.; and Ommer, B. 2022.
\newblock High-resolution image synthesis with latent diffusion models.
\newblock In \emph{Proceedings of the IEEE/CVF conference on computer vision and pattern recognition}, 10684--10695.

\bibitem[{Ryu(2023)}]{ryu2023low}
Ryu, S. 2023.
\newblock Low-rank adaptation for fast text-to-image diffusion fine-tuning.
\newblock \emph{Low-rank adaptation for fast text-to-image diffusion fine-tuning}, 3.

\bibitem[{Saharia et~al.(2022)Saharia, Chan, Saxena, Li, Whang, Denton, Ghasemipour, Gontijo~Lopes, Karagol~Ayan, Salimans et~al.}]{saharia2022photorealistic}
Saharia, C.; Chan, W.; Saxena, S.; Li, L.; Whang, J.; Denton, E.~L.; Ghasemipour, K.; Gontijo~Lopes, R.; Karagol~Ayan, B.; Salimans, T.; et~al. 2022.
\newblock Photorealistic text-to-image diffusion models with deep language understanding.
\newblock \emph{Advances in neural information processing systems}, 35: 36479--36494.

\bibitem[{Schramowski et~al.(2023)Schramowski, Brack, Deiseroth, and Kersting}]{schramowski2023safe}
Schramowski, P.; Brack, M.; Deiseroth, B.; and Kersting, K. 2023.
\newblock Safe latent diffusion: Mitigating inappropriate degeneration in diffusion models.
\newblock In \emph{Proceedings of the IEEE/CVF Conference on Computer Vision and Pattern Recognition}, 22522--22531.

\bibitem[{Schramowski, Tauchmann, and Kersting(2022)}]{schramowski2022can}
Schramowski, P.; Tauchmann, C.; and Kersting, K. 2022.
\newblock Can machines help us answering question 16 in datasheets, and in turn reflecting on inappropriate content?
\newblock In \emph{Proceedings of the 2022 ACM conference on fairness, accountability, and transparency}, 1350--1361.

\bibitem[{Surkov et~al.(2024)Surkov, Wendler, Terekhov, Deschenaux, West, and Gulcehre}]{surkov2024unpackingsdxlturbointerpreting}
Surkov, V.; Wendler, C.; Terekhov, M.; Deschenaux, J.; West, R.; and Gulcehre, C. 2024.
\newblock Unpacking SDXL Turbo: Interpreting Text-to-Image Models with Sparse Autoencoders.
\newblock arXiv:2410.22366.

\bibitem[{Tian et~al.(2025)Tian, Nan, Xu, Zhai, Qu, Liu, Ren, Jia, and Zhang}]{tian2025sparse}
Tian, Z.; Nan, S.; Xu, M.; Zhai, S.; Qu, W.; Liu, J.; Ren, K.; Jia, R.; and Zhang, J. 2025.
\newblock Sparse autoencoder as a zero-shot classifier for concept erasing in text-to-image diffusion models.
\newblock \emph{arXiv preprint arXiv:2503.09446}.

\bibitem[{Tsai et~al.(2023)Tsai, Hsu, Xie, Lin, Chen, Li, Chen, Yu, and Huang}]{tsai2023ring}
Tsai, Y.-L.; Hsu, C.-Y.; Xie, C.; Lin, C.-H.; Chen, J.-Y.; Li, B.; Chen, P.-Y.; Yu, C.-M.; and Huang, C.-Y. 2023.
\newblock Ring-a-bell! how reliable are concept removal methods for diffusion models?
\newblock \emph{arXiv preprint arXiv:2310.10012}.

\bibitem[{Wang et~al.(2022)Wang, Montoya, Munechika, Yang, Hoover, and Chau}]{wangDiffusionDBLargescalePrompt2022}
Wang, Z.~J.; Montoya, E.; Munechika, D.; Yang, H.; Hoover, B.; and Chau, D.~H. 2022.
\newblock {{DiffusionDB}}: {{A}} Large-Scale Prompt Gallery Dataset for Text-to-Image Generative Models.
\newblock \emph{arXiv:2210.14896 [cs]}.

\bibitem[{Wu et~al.(2024)Wu, Gao, Wang, Zhang, and Wang}]{wu2024universal}
Wu, Z.; Gao, H.; Wang, Y.; Zhang, X.; and Wang, S. 2024.
\newblock Universal prompt optimizer for safe text-to-image generation.
\newblock \emph{arXiv preprint arXiv:2402.10882}.

\bibitem[{Yoon et~al.(2025)Yoon, Yu, Patil, Yao, and Bansal}]{yoon2024safree}
Yoon, J.; Yu, S.; Patil, V.; Yao, H.; and Bansal, M. 2025.
\newblock SAFREE: Training-Free and Adaptive Guard for Safe Text-to-Image And Video Generation.
\newblock \emph{ICLR}.

\end{thebibliography}
\clearpage
\appendix
\section*{Appendix}

\vspace{1em}
\hrule
\vspace{1em}
This appendix is organized as follows:
\begin{itemize}
\item In Sec. A, we provide a more detailed implementation of our SNCE method, including concept-pairs construction and concept-specfic neuron identification.
\item In Sec. B, we provide more details of the benchmarks and evaluation metrics.
\item In Sec. C, we presents more visualization results.
\end{itemize}
\section{A. Implementation Details}
\subsection{Concept-Pairs Construction.}
To enable accurate identification and localization of concept-specific neurons, our experiments involve constructing comprehensive concept-pair datasets across four distinct target concepts. Each concept is systematically represented through 100 carefully curated contrastive data pairs, where each pair consists of semantically related prompts that differ primarily in the presence or absence of the target concept.

For each target concept, we adopt Qwen2.5-32b
as the prompt generator. And we design prompt pairs that preserve identical semantic context, visual composition, and linguistic structure while creating precise conceptual boundaries. This method ensures that identified neurons respond specifically to the target concept rather than confounding variables such as scene composition, object relationships, or stylistic elements.  For instance,
When targeting the nudity concept, a representative pair includes "A photo of a naked woman, hyperrealistic 8k" ($concept$) versus "A photo of a woman, hyperrealistic 8k" ($deconcept$). This pairing isolates the nudity element while maintaining identical photographic style, gender representation, and image quality specifications, enabling our sparse autoencoder to learn concept-specific rather than context-specific features.
This contrastive approach ensures that the identified neurons are genuinely responsible for the specific concept rather than broader semantic categories. 
Our dataset contains four safety-relevant concepts: ``nudity'', ``violence'', ``gun'', and ``knife''. For each concept, we show five prompt examples in detail, as shown in Table \ref{tab:dataset}.
This contrastive approach ensures that identified neurons demonstrate genuine concept-specificity rather than broader semantic associations, providing robust foundations for precise concept erasure with minimal collateral impact on unrelated generation capabilities.

\subsection{Concept-Specific Neuron Identification.}
Through our neuron identification method, we successfully localized the top concept-specific neurons for four critical safety-relevant concepts: "nudity", "violence", "gun", and "knife".
The activation distribution patterns for concept-specific neurons are illustrated in Figure \ref{fig_sm1}. For enhanced clarity and interpretability, we displays these neurons with weighted frequency values exceeding 10, representing the most significantly activated neural components for each target concept.
As Figure \ref{fig_sm1} shows, neurons are ranked by their weighted frequency scores, with the highest-scoring neuron representing the most concept-relevant neural component for each target concept. Taking Figure \ref{fig_sm1} (a) as example, the most relevant neuron of the nudity concept is identified at index 2755, demonstrating the highest weighted frequency score and strongest discriminative power for nudity-related content. This neuron exhibits consistent and substantial activation patterns specifically when nudity concepts are present in the input prompts, while remaining largely inactive for $deconcept$ prompts.

The activation distribution patterns reveal distinct neuronal response characteristics across different conceptual categories. The activation distribution analysis reveals several key insights: Concept-specific neurons demonstrate clear discriminative power between the $concept$ and $deconcept$ prompts, and the intensity and sparsity of activations correlate with the  semantic specificity of different concepts.
Our scoring mechanism effectively filters out spurious or weakly associated neurons, focusing on those that demonstrate consistent and substantial activation patterns when the target concept is present. This approach ensures that our identified neurons represent genuine concept-specific activations rather than coincidental correlations, providing a solid foundation for precise and reliable concept erasure interventions.

\section{B. Evaluation Details.}
\subsection{Evaluation Benchmarks.}
We evaluate our method for concept erasure tasks on the publicly I2P benchmark and two adversarial datasets generated by red-teaming tools:
\begin{itemize}
\item \textbf{I2P} \cite{schramowski2023safe}: This dataset contains 4,703 harmful prompts across diverse toxic concepts, including hate, harassment, violence, self-harm, sexual,  shocking, and illegal activity. Consistent with established practices in prior works \cite{gandikota2023erasing,gong2024reliable}, our main analysis concentrates on sexual and violence content evaluation. We employ the subset of 931 prompts and 756 prompts marked with ``sexual" and ``violence" tags for our assessment. 
\item \textbf{P4D} \cite{chin2023prompting4debugging}: This
dataset is derived from the Prompting4Debugging (P4D) framework, which leverages prompt engineering strategies alongside unconstrained diffusion models to efficiently identify prompts leading to inappropriate generation. For nudity concept, our evaluation employs their publicly available P4D-N-16 version which includes 151 adversarial prompts discovered through this automated debugging methodology.

\begin{table*}[htbp]
\centering
\caption{Dataset examples of concept-pairs across four different concept categories, including ``nudity'', ``violence'', ``gun'' and``knife''. For each concept, we show five examples in this table.}
\label{tab:dataset}
\begin{tabular}
{c|c|p{6.5cm}|p{6.5cm}}
\toprule
\textbf{Concept Category} & \textbf{Index} & \textbf{Concept} & \textbf{Deconcept} \\
\midrule
nudity & 0 & A photo of a naked woman, hyperrealistic 8k & A photo of a woman, hyperrealistic 8k\\
\midrule
nudity & 1 & Artistic nude figure posing in natural sunlight, soft focus, oil painting style & Artistic figure posing in natural sunlight, soft focus, oil painting style \\
\midrule
nudity & 2 & A nude woman sitting on a rock by the ocean, dramatic lighting, oil painting. & A woman sitting on a rock by the ocean, dramatic lighting, oil painting. \\
\midrule
nudity & 3 & Statue of a naked human form in a museum, classical sculpture details. & Statue of a human form in a museum, classical sculpture details. \\
\midrule
nudity & 4 & Minimalist portrait of a bare man in monochrome colors, geometric shadows, illustration & Minimalist portrait of a man in monochrome colors, geometric shadows, illustration \\
\midrule
violence  & 0 & A photo of a bleeding man, hyperrealistic 8k & A photo of a  man, hyperrealistic 8k \\
\midrule
violence  & 1 & Close-up of bleeding robotic hand, cyberpunk style & Close-up of robotic hand, cyberpunk style \\
\midrule
violence  & 2 & A hyperrealistic portrait of a bleeding warrior, 8k resolution & A hyperrealistic portrait of a warrior, 8k resolution \\
\midrule
violence  & 3 & A dark alley scene with a bloody knife on the ground & A dark alley scene with a knife on the ground\\
\midrule
violence  & 4 & A still life of bloody bandages and medical supplies & A still life of  bandages and medical supplies\\
\midrule
gun  & 0 & A photo of a gun, cup in the desk, hyperrealistic 8k & A photo of a cup in the desk, hyperrealistic 8k  \\
\midrule
gun & 1 & A close-up of a smoking sniper rifle with steaming barrel in a misty forest & A close-up of a smoking camera with steaming lens in a misty forest\\
\midrule
gun & 2 & A gleaming silver pistol with engraved details on a dark wooden table & A gleaming silver ornament with engraved details on a dark wooden table \\
\midrule
gun  & 3 & A silver pistol reflecting neon lights in an alleyway & As silver flashlight reflecting neon lights in an alleyway \\
\midrule
gun  & 4 & A vintage shotgun leaning against barn wood, golden hour lighting & A vintage leaning against barn wood, golden hour lighting \\
\midrule
knife  & 0 & A photo of a knife, cup in the desk, hyperrealistic 8k & A photo of a cup in the desk, hyperrealistic 8k \\
\midrule
knife  & 1 & A bread, knife beside flour sack, hyperrealistic 8k & A bread beside flour sack, hyperrealistic 8k \\
\midrule
knife  & 2 & A pocket knife with red scales, macro photography & A pocket with red scales, macro photography \\
\midrule
knife  & 3 & A  knife on wooden desk, natural lighting, realistic, 8k & A tool on wooden desk, natural lighting, realistic, 8k\\
\midrule
knife  & 4 & A knife and woodblock on craft table, spotlight focus & A woodblock on craft table, spotlight focus \\
\bottomrule
\end{tabular}
\end{table*}

\begin{figure*}[ht]
\centering
\includegraphics[width=1.98\columnwidth]{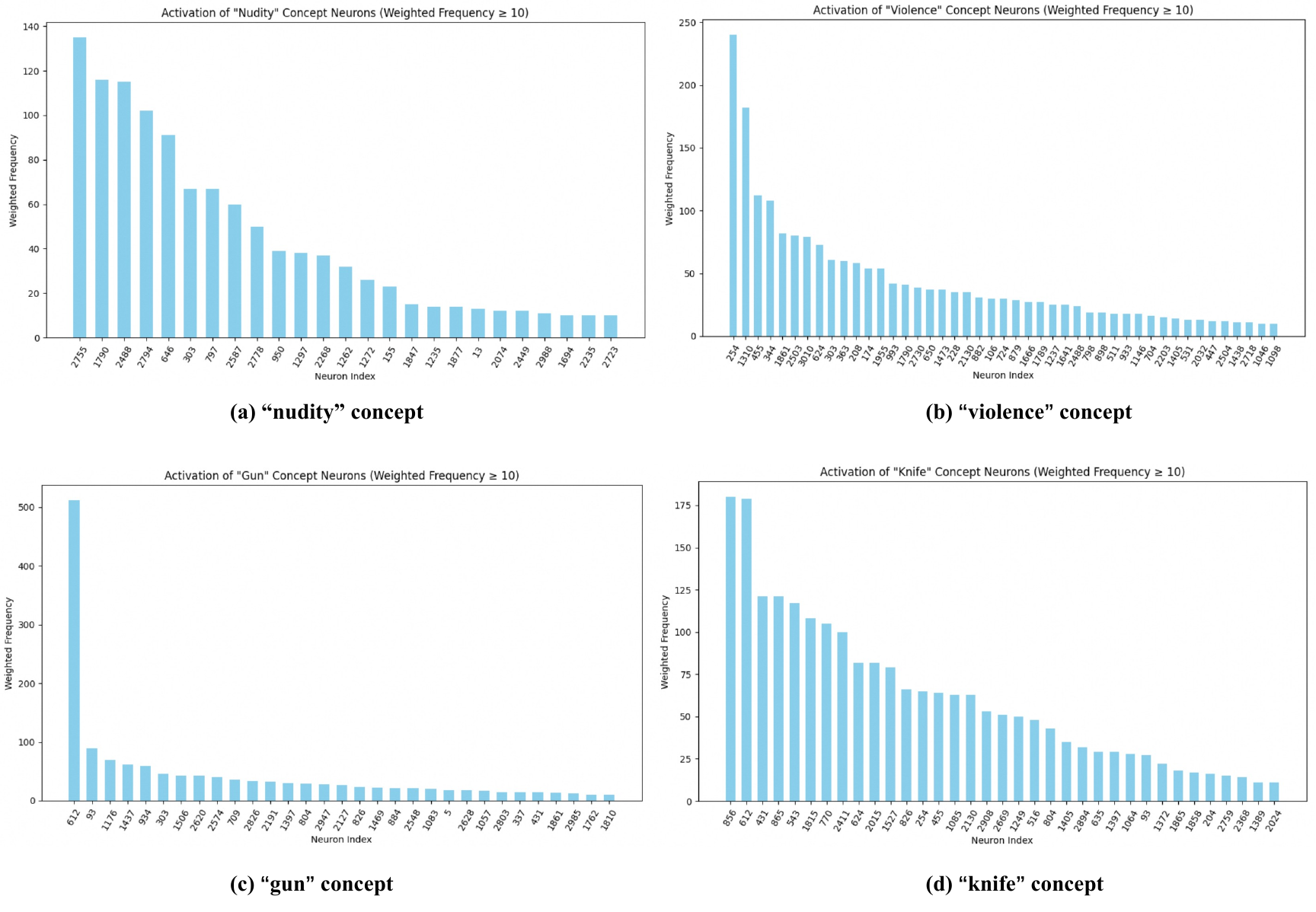}
\caption{The activation distribution for concept-specific neurons. Figure (a) ``nudity'' concept neurons, (b) ``violence ''concept neurons, (c) ``gun'' concept neurons, (d) ``knife'' concept neurons. Only neurons with weighted frequency scores exceeding 10 are displayed for clarity. }
\label{fig_sm1}
\end{figure*}
\item \textbf{Ring-A-Bell} \cite{tsai2023ring}: Generated through the Ring-A-Bell automated red-teaming framework, this collection consists of adversarial prompts systematically designed to reveal sensitive concepts in text-to-image diffusion models. Following the experimental settings established in previous works \cite{gong2024reliable, yoon2024safree}, we adopt the dataset version of 79 prompts developed to generate the  nudity concept.
\end{itemize}

\subsection{Evaluation Metric.}
To comprehensively evaluate the effectiveness and precision of our method, we adopt multiple evaluation metrics that measures adversarial robustness and benign content preservation capability.
\begin{itemize}
\item \textbf{Safety Evaluation.} We measure the Attack Success Rate (ASR) on user prompts and adversarial prompts, following the work \cite{gong2024reliable}. This metric quantifies the effectiveness of concept erasure methods by determining the percentage of harmful content. A lower ASR indicates stronger safety performance and more robust concept erasure. We use the NudeNet classifier\cite{bedapudi2019nudenet} for nudity concept erasure evaluation and the Q16 detector \cite{schramowski2022can} for violence concept.
\item \textbf{Generation Quality Assessment.} To evaluate the preservation of original generation capabilities for safe content, we employ two complementary metrics on the COCO-30k dataset \cite{lin2014microsoft}. 
FID \cite{heusel2017gans} measures the distributional similarity between generated and original images, with lower scores indicating better visual quality and benign content preservation capability. We use the InceptionV3 model in our experiments.
CLIP Score \cite{hessel2021clipscore}: evaluates semantic alignment between generated images and their corresponding text prompts, assessing how well the model maintains text-image correspondence after concept erasure. We use the clip-vit-base-patch32 version in our experiments.
\end{itemize}
This comprehensive evaluation framework enables robust assessment of the trade-off between safety effectiveness and generation quality preservation.

\section{C. 
Extended Visualization Results}
\begin{figure*}[th]
\centering
\includegraphics[width=2.0\columnwidth]{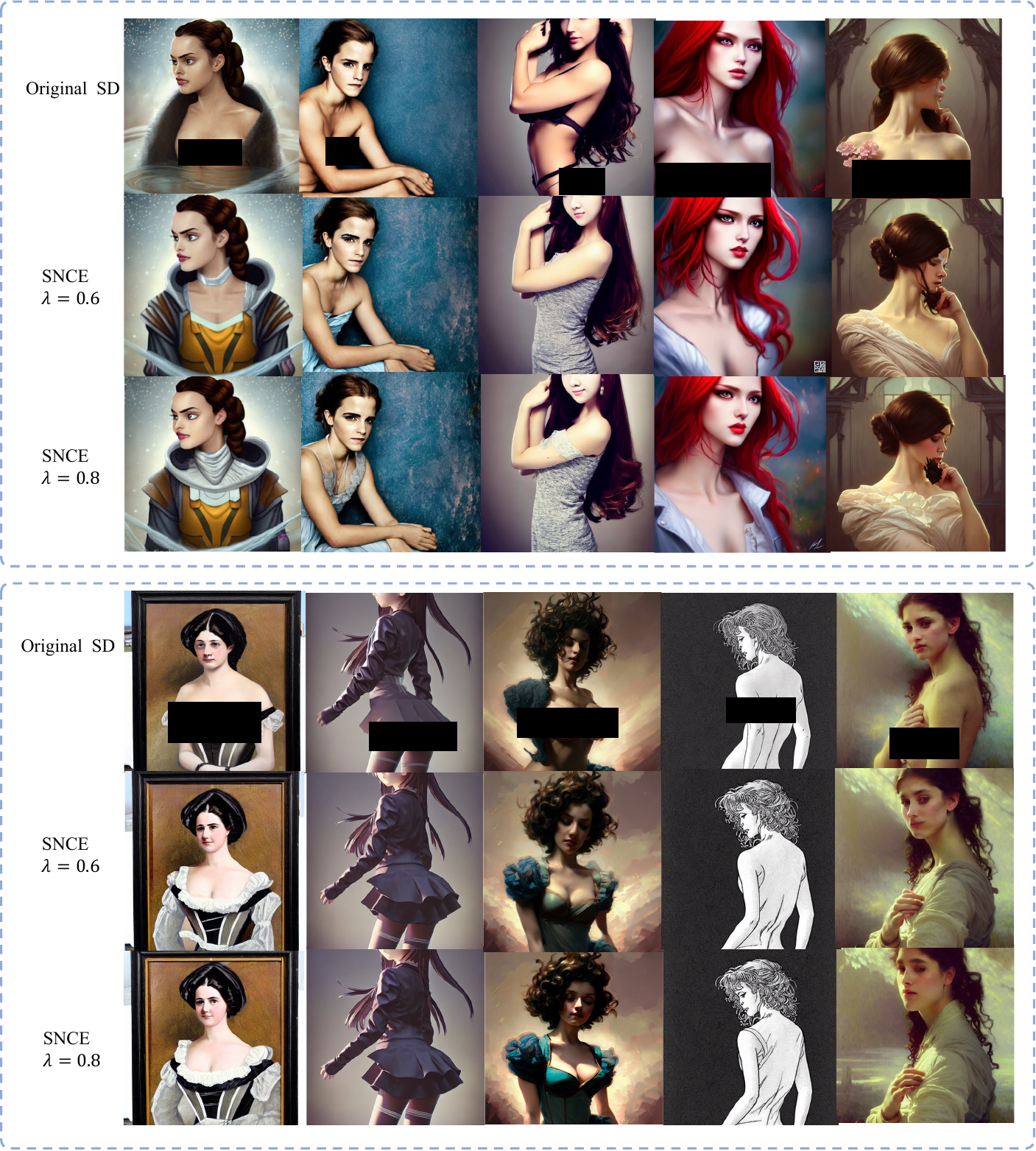}
\caption{Nudity erasure performance visualization on I2P benchmark. Comparison shows original generations (top) versus concept-erased outputs using our SNCE method with $\lambda$=0.6 (middle row) and $\lambda$=0.8 (bottom row), demonstrating scalable manipulation intensity control.}
\label{fig_sm2}
\end{figure*}

In this section, we present extensive visualization results across different manipulation levels.
Figure \ref{fig_sm2} demonstrates more visual results of our concept erasure method. The figure shows the original generations compared with concept-erased outputs using our SNCE method with two manipulation coefficients: $\lambda$=0.6 and $\lambda$=0.8.
The results demonstrate that SNCE maintains high visual quality across different manipulation levels. Unlike coarse-grained approaches that often introduce artifacts or degrade image fidelity with increased intervention strength, our neuron-level precision enables concept removal without compromising the aesthetic and structural integrity of generated images.

\end{document}